\documentclass[10pt,a4paper]{article}
\usepackage[utf8]{inputenc}
\usepackage{amsmath}
\usepackage{amsfonts}
\usepackage{amssymb}
\usepackage{apacite}
\usepackage{color}
\usepackage{graphicx}
\usepackage{subcaption}
\usepackage[linesnumbered,ruled,vlined]{algorithm2e}
\usepackage{geometry}
\title{Tweet Sentiment Extraction using Viterbi Algorithm with Transfer Learning}
\def\correspondingauthor{\footnote{Corresponding author: ziedbaklouti@outlook.com}}
\def\university{\footnote{PARIS CITÉ UNIVERSITY}}
\author{Zied Baklouti\correspondingauthor{} \university{}}
\date{August 2023}

\begin{document}
\maketitle

	\begin{abstract}

\begin{flushleft}
Tweet sentiment extraction extracts the most significant portion of the sentence, determining whether the sentiment is positive or negative. This research aims to identify the part of tweet sentences that strikes any emotion.
To reach this objective, we continue improving the Viterbi algorithm previously modified by the author to make it able to receive pre-trained model parameters. We introduce the confidence score and vector as two indicators responsible for evaluating the model internally before assessing the final results. We then present a method to fine-tune this nonparametric model. We found that the model gets highly explainable as the confidence score vector reveals precisely where the least confidence predicted states are and if the modifications approved ameliorate the confidence score or if the tuning is going in the wrong direction.
\end{flushleft}

	\end{abstract}
	
	\section{Introduction}
	
		Determining the sentiment of a tweet can be a laborious task for NLP specialists, as they need to identify the specific segment of the sentence that accurately reflects the sentiment and its boundaries. It can be challenging to accomplish this task when the sentences are lengthy and the intended emotion is conveyed using multiple words or placed at the start or end. 
 
	Information extraction and sentiment analysis are indispensable for processing news feeds and posts from public profiles of celebrities and ordinary persons to determine the sentiment of a tweet. When automated, these activities allow the categorization of tweets into several predefined classes and perhaps avoid the diffusion of fake news or toxic posts. Emotional writing can engage users and encourage them to spend more time browsing a website or getting more information about a product. However, it can also negatively impact the reader's mood, especially when they come across a toxic text with a high frequency of negative emotions, such as insulting comments or discriminatory remarks from followers on social media. Detecting such infractions early can increase the audience number on a web page and avoid unsubscribing clicks. 
	
When it comes to opinion mining, analyzing public opinion can be highly beneficial in assessing satisfaction and agreement with political decisions and programs. This type of analysis can offer valuable insights into a candidate's popularity and even aid in predicting their likelihood of winning an election compared to their competitors. {\color{blue}\cite{10.1145/3442442.3452322}}{\color{blue}\cite{9679946}}.

In machine translation systems, Identifying a sentence appearing sentiment can also help traduction systems evaluate the correct meaning generated by a token, develop a traduced text with high accuracy, and keep the original text sentiments and nuances {\color{blue}\cite{xu2018unpaired}}{\color{blue}\cite{mohammad2016translation}}. 

By gathering data through keywords, marketing agencies can determine whether their product advertisements effectively reach the intended audience and whether that audience is engaging with the posts{\color{blue}\cite{rambocas2018online}}. Additionally, analyzing comments and posts on public profiles can provide insight into the interests of a particular group, allowing targeted advertising to reach individuals with similar hobbies or areas of focus{\color{blue}\cite{fan2010sentiment}}{\color{blue}\cite{qiu2010dasa}}.

Twitter is an excellent platform for extracting sentiment since various users from diverse fields express their opinions or announce upcoming events through tweets or textual posts. Since tweets could be considered a sequence of words, we can use an approved NLP model to perform this task. Still, this article aims to develop a new model using the transfer learning capabilities of transferring pre-trained model parameters. We used a portion of the tweet sentiment extraction dataset{\color{blue}\cite{tweet-sentiment-extraction}}. The dataset and the Matlab scripts used are available from this GitHub link: $https://github.com/Zied130/Tweet\_Sentiment$- . 
	
	\section{Related Work}
	
The Viterbi algorithm is a dynamic programming algorithm used in various scientific models to predict the most probable sequence of hidden states in a Hidden Markov Model (HMM). In NLP, HMM models are primarily used to determine sequences of part-of-speech (POS) tagging{\color{blue}\cite{toutanvoa-manning-2000-enriching}}{\color{blue}\cite{toutanova-etal-2003-feature}}, named entity recognition (NER){\color{blue}\cite{ratinov-roth-2009-design}}, or speech recognition {\color{blue}\cite{18626}}. 
	
Contextualizing information is an important aspect of information extraction. The words and POS that follow a predicted state can give insight into the status of the current token in an HMM model. {\color{blue}\cite{ratinov-roth-2009-design}}. 

To address the issue of interpretability in new NLP models and the challenges educators and learners face in comprehending the workings of large language models, experts in NLP are placing significant emphasis on creating explainable models{\color{blue}\cite{10.1016/j.inffus.2019.12.012}}. They aim to achieve a level of intelligence comparable to humans using limited memorized observations. 

When designing a suitable model for NLP tasks, algorithm concepts rely on interdisciplinarity as an essential criterion. To develop more robust models, it is necessary to understand the theory behind the decision-making procedure of information extraction{\color{blue}\cite{Baklouti2019HiddenMB}}{\color{blue}\cite{Baklouti2021ExternalKT}}. Using punctuation when writing textual data, especially for long sentences, is crucial. It serves as an excellent indicator of how the human brain needs to restructure long structures into smaller ones to assimilate information better. This helps to avoid losing attention to the words' meaning and combinations.

Transfer learning is a technique that enhances the capability of NLP models to carry out information extraction tasks. It achieves this by improving the model's interpretability and combining contextual information in categorical or quantified form with the predefined model. This is achieved by transferring knowledge from another related model{\color{blue}\cite{Baklouti2021ExternalKT}}.

Getting the best performance out of deep neural network models demands a lot of computation power because these models have intricate architectures comprising various layers and parameters. It's not straightforward to grasp how this specific model operates. To ensure explainable AI, it is essential to have transparency and avoid relying solely on black-box models{\color{blue}\cite{10.1145/3236009}}{\color{blue}\cite{rudin2019stop}}. 

Labeling the inputs is also an essential procedure for AI-based systems, and for more interpretable mechanisms, observing a strong dependence between the model structure sizes and the labeling criteria is an excellent step toward an explainable AI. For DNN models, the choice of nodes and layer number doesn't variate during the model implementation leading to static interpretation of the model performances, and the improvements are based on more hyperparameters fine-tuning {\color{blue}\cite{10.1145/3236386.3241340}}. 

\section{Methodology}
We used the last advances in the Viterbi algorithm developed by the author to incorporate external knowledge into this algorithm logic{\color{blue}\cite{Baklouti2021ExternalKT}}. There are three agents used in our model to extract the sentiment in a tweet:
\begin{itemize}
\item Word-level tokens: We utilized the 'tokenizedDocument' function from the Text Analytics Toolbox of Matlab to extract tokens from the processed text. The phrase-level tokens in the output were formed solely with words from the Word-level tokens generated.

\item POS tags: We attributed to each token a POS tag generated by the function 'addPartOfSpeechDetails' of the Text Analytics Toolbox of Matlab. This agent represents an external transferred knowledge to the HMM model.

\item coeff: This is an estimation of the coefficient for a generalized linear model that predicts the sentiment category of a sentence as either negative, neutral, or positive. The input data for this model is generated using the 'bagOfWords' function from the Text Analytics Toolbox of Matlab. This agent represents an external transferred knowledge to the HMM model.
\end{itemize}

We used the Viterbi algorithm with transfer learning to extract tweet sentiment. This involved a modified version of the double-agent Viterbi algorithm. The parameters of the HMM model used were as follows:

\begin{itemize}
\item $s \in \{0,1,2,3,...\}$, the states of the model, if the token is not selected then $s=0$ else if the token is selected $s>0$. For selected tokens $s=1$ for the first token and the other states depend on the POS tag of the token, if the $POS \in \{adposition, punctuation, coord-conjunction, pronoun, particle, subord\-conjunction\}$ then $s=1$ automatically and else the state is incremented by one $s_{r}=s_{r-1}+1$  
\item $A=\{ a_{coeff, POS, i,k} \}$, the state transition probability distribution. The probability $a_{coeff, POS, i,k}$ is the probability that the system will move in one transition from state $i$ to state $k$ given at state $k$ POS and coeff agents are known.   
\item $B=\{ b_{POS, i,k,V} \}$, the observation symbol probability distribution. The probability $b_{POS, i,k, V}$ is the probability that the observation $V$ is emitted in position $r$ when the model moves from state $i$ in position $r$ to state $k$ in position $r+1$ given at state $k$ POS is known.
\item $A_I=\{ \pi_{coeff, i}^{a} \}$, the initial state distribution. The probability $\pi_{coeff, i}^{a}$ is the probability that the system will start in state $i$ given that the initial coeff is known.   
\item $B_I=\{ \pi_{i,POS,V}^{b} \}$, the initial observation symbol probability distribution. The probability $\pi_{i,POS,V}^{b}$ is the probability that the observation $V$ will be emitted at the beginning of the sentence when the model has the initial state $i$  given at the initial state $i$ POS is known.
\item $C_I=\{ \pi_{POS, i,k,V}^{c} \}$, the initial observation symbol probability distribution with initial states transition. The probability $\pi_{POS, i,k,V}^{c}$ is the probability that the token $V$ will be emitted at the beginning of the sentence when the model moves from the initial state $i$ to the second state $k$  given at state $i$ POS is known.
\end{itemize}

To evaluate the model, two scores are used: the Jaccard score and the confidence score. The confidence score is utilized to determine the model's level of certainty in estimating the output's state. If the confidence score is one, the model is highly confident in its estimation, and there's a high likelihood of obtaining an accurate estimation. However, if the confidence score decreases, there's a lower level of confidence, which leads to hesitation regarding the estimation. The confidence score is calculated using the following formulas: 

\begin{gather} 
\text{confidence score} = \frac{\sum (\text{confidence vector} \times w^{'})}{ \sum_l w_l } \label{CS}
 \end{gather}

\begin{gather*} 
		\text{where  } w_l = \begin{cases}
	0 &  token_i \text{ is not selected regarding the output vector}  \\
	\,  \\
	1 &  token_i \text{ is selected regarding the output vector} \\
	\end{cases}\\
	l = 1,2,...,\text{sequence length }(L)
		 \end{gather*}
		 
	\begin{gather}	 
	\text{confidence vector}[k] = \begin{cases}
	\frac{Delta_{L-1}(output(L-1),output(L))}{\sum_j Delta_{L-1}(output(L-1),j)} &   \text{ for k = L}  \\  
	\frac{Delta_{L-1}(output(L-1),output(L))}{\sum_i Delta_{L-1}(i,output(L))} &   \text{ for k = L-1} \\
	CM_{L-2}(output(L-1),output(L)) &   \text{ for k = 1,2,..,L-2} 
	\end{cases}  \label{CV}
\end{gather}

Where $CM$ is a set of confidence matrices obtained using the formulas : 

\begin{gather}
confidence\_matrix(k,j)= \frac{max_i(Delta_{l-1}(i,k) \times v_i)}{\sum_i (Delta_{l-1}(i,k) \times v_i)} \label{CM}
\end{gather}

\begin{gather}
 v_i=P(\Gamma_r=k|\Gamma_{r-1}=i,POS_l,a_l<=coeff_l<b_l) \label{VI}
\end{gather}

Where $i = 1,2,...,rows(Delta_{l-1})$ , $k = 1,2,...,rows(Delta_{l})$ and $j = 1,2,...,columns(Delta_{l})$.

\begin{gather*} 
		\text{where  } [a_l,b_l] = \begin{cases}
	[round(coeff_l),round(coeff_l) + \text{sample rate} ] &  \text{if }round(coeff_l) <= coeff_l  \\
	\,  \\
	[round(coeff_l) - \text{sample rate} , round(coeff_l) ] &  \text{if }round(coeff_l) > coeff_l \\
	\end{cases}
		 \end{gather*}

The confidence score equation is represented by Equation {\color{blue}\eqref{CS}}. The vector $w$ has the same length as the output vector, which contains the states $s_l$ (where $l=1,2...L$). Additionally, Equation {\color{blue}\eqref{CV}} represents the confidence vector equation, which contains the confidence score of each token. Equation {\color{blue}\eqref{CM}} represents confidence matrix used to backtrack confidencce scores. Equation {\color{blue}\eqref{VI}} represents the model parameter $A=\{ a_{coeff, POS, i,k} \}$.

\begin{algorithm}[H]
\DontPrintSemicolon
\tcc{Initialization step}
\If{$POS_2 \in $ \{'adposition','punctuation','coord-conjunction','pronoun','particle','subord-conjunction'\} }
{
$\delta=zeros(2,2)$ \\
\For{$i=1:2$}{\For{$j=1:2$}{$\delta(i,j)= P(\Gamma_1=i|token\_number=1,a_1<=coeff_1<b_1) \times P(V_1|POS_1,\Gamma_1=i,token\_number=1) \times P(V_1|\Gamma_1=i,\Gamma_2=j,POS_1,token\_number=1)$}}}
\Else
{
$\delta=zeros(2,3)$ \\
\For{$i=1:2$}{\For{$j=1:3$}{$\delta(i,j)= P(\Gamma_1=i|token\_number=1,a_1<=coeff_1<b_1) \times P(V_1|POS_1,\Gamma_1=i,token\_number=1) \times P(V_1|\Gamma_1=i,\Gamma_2=j,POS_1,token\_number=1)$}}}
$Delta_1 \leftarrow Normalize(\delta) $ \\
\tcc{Iteration step}
$L=sequence\_length$ \\
\For{l=2:$L -1$ }{\If{$POS_{l+1} \in $ \{'adposition','punctuation','coord-conjunction','pronoun','particle','subord-conjunction'\} }
{rows=columns($Delta_{l-1}$) \\
columns=2
}
\Else{
rows=columns($Delta_{l-1}$) \\
columns=columns($Delta_{l-1}$) + 1
}
$\delta=zeros(rows,columns)$ \\
$\psi=zeros(rows,columns)$ \\
$confidence\_matrix=zeros(rows,columns)$ \\
\For{k=1:rows}
{
$v=zeros(rows(Delta_{l-1}))$ \\
\For{i=1:rows($Delta_{l-1}$)}
{
$v_i=P(\Gamma_r=k|\Gamma_{r-1}=i,POS_l,a_l<=coeff_l<b_l)$
}
\For{j=1:columns}
{
$\delta(k,j) = max_i(Delta_{l-1}(i,k) \times v_i) \times P(V_l|\Gamma_r=j,\Gamma_{r-1}=k,POS_l)$ \\
$\psi(k,j)= argmax_i(Delta_{l-1}(i,k) \times v_i)$ \\
$ confidence\_matrix(k,j)= \frac{max_i(Delta_{l-1}(i,k) \times v_i)}{\sum_i (Delta_{l-1}(i,k) \times v_i)} $ \\
}
}
$Delta_l \leftarrow Normalize(\delta) $ \\
$\Psi_{l-1} \leftarrow \psi $ \\
$CM_{l-1} \leftarrow confidence\_matrix $ \\
}
\tcc{Termination step}
$output=zeros(L)$ \\
$output(L)= arg_j max_{1<=i<=rows(Delta_{L-1}),1<=j<=columns(Delta_{L-1})} Delta_{L-1}(i,j) $ \\
$output(L-1)= arg_i max_{1<=i<=rows(Delta_{L-1}),1<=j<=columns(Delta_{L-1})} Delta_{L-1}(i,j) $ \\
$output(L-2)= \Psi_{L-2}(output(L-1),output(L)) $ \\
$confidence\_vector=zeros(L)$ \\
$confidence\_vector(L)=\frac{Delta_{L-1}(output(L-1),output(L))}{\sum_j Delta_{L-1}(output(L-1),j)}$ \\
$confidence\_vector(L-1)=\frac{Delta_{L-1}(output(L-1),output(L))}{\sum_i Delta_{L-1}(i,output(L))}$ \\
$confidence\_vector(L-2)=CM_{L-2}(output(L-1),output(L))$ \\
\tcc{Backtracking step} 
\For{i=L-3:1}
{
$output(i)=\Psi_i(output(i+1),output(i+2))$ \\
$confidence\_vector(i)=CM_i(output(i+1),output(i+2))$
}
\caption{Viterbi With Transfer Learning}
\end{algorithm}

\section{Results and Interpretation }

The table{\color{blue}[\ref{table:table1}]} shows a clear connection between the confidence and Jaccardi scores. As the confidence score goes up, so does the Jaccardi score. This is because the confidence score reveals whether a sentiment exists in a specific position within a sentence and indicates how sure the system is about extracting that particular token as a part of the words that make up the sentiment. This metric improves the Jaccardi score by addressing the extracted emotion's boundaries and preventing unwanted tokens' extraction.

\begin{table}[h]
\center
\begin{tabular}{|c|c|c|}
\hline 
Viterbi Form number  & Jaccard training score & Confidence training score \\ 
\hline 
viterbi Form One  & 0.9485 & 0.9825 \\ 
\hline 
viterbi Form Two  & 0.9681 & 0.991 \\ 
\hline 
viterbi Form Three & 0.9746 & 0.996 \\ 
\hline 
\end{tabular}
\caption{Jaccard score and confidence score for the three Viterbi Forms}
\label{table:table1}
\end{table} 

\begin{figure}[h]
\centering
\begin{subfigure}[t]{0.5 \textwidth}
\centering
\includegraphics[height=2in]{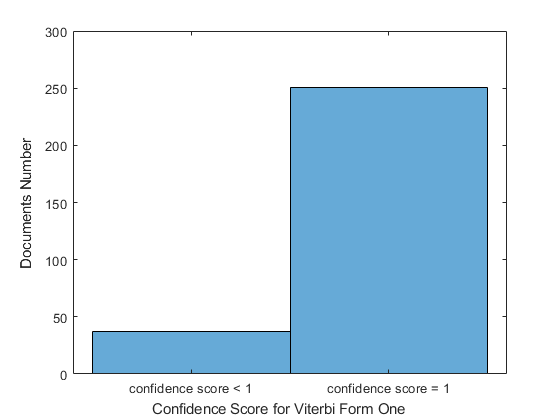}
\caption{}
\end{subfigure}%
~
\begin{subfigure}[t]{0.5 \textwidth}
\centering
\includegraphics[height=2in]{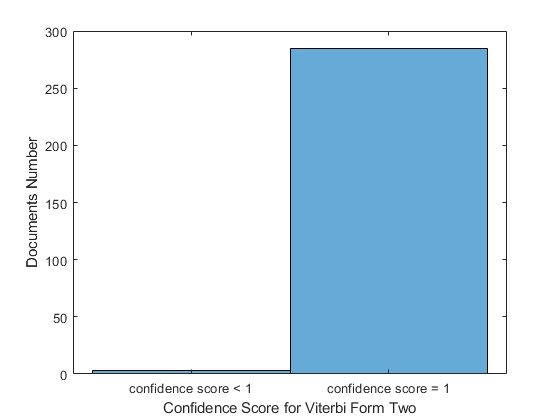}
\caption{}
\end{subfigure}%
\caption{Value distribution of the last value in the confidence score vectors using Viterbi Form one and Viterbi Form two}
\label{fig:fig1}
\end{figure}

Additionally, the confidence score can identify areas where the model needs improvement by incorporating additional parameters. In Figure {\color{blue}[\ref{fig:fig1}]}, we compare the confidence score values for the last tokens in sentences between Viterbi Form One presented by equation {\color{blue}\eqref{EQ1}} and Viterbi Form Two presented by equation {\color{blue}\eqref{EQ2}}, which includes an extra parameter. This parameter, $C=\{ c_{POS, k,j,V_{+1}} \}$, is the probability distribution of the delayed observation symbol. The probability $c_{POS, k,j, V_{+1}}$ refers to the probability that observation $V_{+1}$ will be emitted at position $r+1$. When the model moves from state $k$ in position $r-1$ to state $j$ in position $r$, given that the POS is known at state $j$. By including this new model parameter, the number of documents with a maximum confidence score of one for the last token increased by 34 (251 for Viterbi Form One and 285 for Viterbi Form Two) for 288 total document numbers. 

\begin{itemize}
\item Viterbi Form One :
\begin{gather}
\delta(k,j) = max_i(Delta_{l-1}(i,k) \times v_i) \times P(V_l|\Gamma_r=j,\Gamma_{r-1}=k,POS_l)  \label{EQ1}
\end{gather}

\item Viterbi Form Two :
\begin{gather}
\delta(k,j) = \begin{cases}
	max_i(Delta_{l-1}(i,k) \times v_i) \times P(V_l|\Gamma_r=j,\Gamma_{r-1}=k,POS_l) &  l<L-1  \\
	\,  \\
	\begin{split}
	max_i(Delta_{l-1}(i,k) \times v_i) \times P(V_l|\Gamma_r=j,\Gamma_{r-1}=k,POS_l) \\ \times P(V_{l+1}|\Gamma_r=j,\Gamma_{r-1}=k,POS_l) \end{split} &  l=L-1  \label{EQ2} \\
	\end{cases}
\end{gather}
\item Viterbi Form Three :
\begin{gather}
\begin{split}
\delta(k,j) & = max_i(Delta_{l-1}(i,k) \times v_i) \times P(V_l|\Gamma_r=j,\Gamma_{r-1}=k,POS_l) \\  & \times P(V_{l+1}|\Gamma_r=j,\Gamma_{r-1}=k,POS_l) \label{EQ3}
\end{split}
\end{gather}

where $l = 1,2,...,L-1$, $k=1,2,...,rows(l)$ , $j=1,2,...,columns(l)$, $Delta$ is the set of state matrix , $\delta$ is the current state matrix, $\Gamma_r$ is state of the output at position $r$ , $i = 1,2,...rows(l-1)$  

\end{itemize}

By adjusting the size of the state matrix in each iteration based on the Pos tags behind the predicted token and the previous state matrix, the interpretability of sentiment extraction models has improved. This adjustment reduces the area of the state matrix around the most likely estimated states, resulting in a more precise estimation and a smaller denominator in the confidence matrix equations {\color{blue}\eqref{CM}}. This leads to a more confident model. 

Improving the model tuning has become easier with a straightforward method for adding new parameters and identifying areas of low confidence to make adjustments. For instance, Viterbi Form One was weak in estimating the last token using equation {\color{blue}\eqref{EQ1}}. Still, by introducing a new model parameter while estimating the state for the last token in equation {\color{blue}\eqref{EQ2}}, we enhanced the model's performance in Viterbi Form Two. This model-tuning approach can also be applied to all tokens in the sentence, as demonstrated in Viterbi Form Three equation {\color{blue}\eqref{EQ3}}. This modification has further increased the model's performance compared to Viterbi Form Two.

\section{Conclusion}

Our paper presents improvements to the Viterbi algorithm, resulting in better performance. We accomplished this by developing a more easily understandable model and incorporating recent advancements through transfer learning techniques. Additionally, we created a state matrix that adapts its dimensions based on contextual information, resulting in a more effective NLP algorithm. We added a conditional probability distribution and calculated the confidence score vector to fine-tune the model. We compared the performances of various Viterbi forms. Our algorithm selected a set of POS tags where the state matrix's second dimension is reset to 2. However, we acknowledge that more research is needed to effectively choose this list of POS tags using a more elaborate method.

\newpage

\bibliographystyle{apacite}
\bibliography{mybib}

\end{document}